\documentclass{sn-jnl}

\usepackage[round, authoryear]{natbib}
\usepackage{graphicx}%
\usepackage{multirow}%
\usepackage{amsmath,amssymb,amsfonts}%
\usepackage{amsthm}%
\usepackage{mathrsfs}%
\usepackage[title]{appendix}%
\usepackage{xcolor}%
\usepackage{textcomp}%
\usepackage{manyfoot}%
\usepackage{booktabs}%
\usepackage{algorithm}%
\usepackage{algorithmicx}%
\usepackage{algpseudocode}%
\usepackage{listings}%
\usepackage{tikz}
\usetikzlibrary{positioning}

\begin{document}

\newcommand\MyBox[1]{%
    \fbox{\parbox[c][1.27cm][c]{1.27cm}{\centering #1}}%
}
\newcommand\MyVBox[1]{%
    \parbox[c][1cm][c]{1cm}{\centering\bfseries #1}%
}  
\newcommand\MyHBox[2][\dimexpr2cm+2\fboxsep\relax]{%
    \parbox[c][1cm][c]{#1}{\centering\bfseries #2}%
}  
\newcommand\MyTBox[4]{%
    \MyVBox{#1}
    \MyBox{#2}\hspace*{-\fboxrule}%
    \MyBox{#3}\par\vspace{-\fboxrule}%
}  
\newcommand*\rot{\rotatebox{90}}

\title[Predicting municipalities in financial distress: a machine learning approach enhanced by domain expertise]{Predicting municipalities in financial distress: a machine learning approach enhanced by domain expertise}


\author[1]{\fnm{Dario} \sur{Piermarini}}\email{dario.piermarini@corteconti.it}
\equalcont{These authors contributed equally to this work.}

\author*[2]{\fnm{Antonio M.} \sur{Sudoso}}\email{antonio.sudoso@iasi.cnr.it}
\equalcont{These authors contributed equally to this work.}

\author[3]{\fnm{Veronica} \sur{Piccialli}}\email{veronica.piccialli@uniroma1.it}

\affil[1]{\orgdiv{Direzione Generale Servizi Informativi Automatizzati}, \orgname{Corte dei Conti}, \orgaddress{\street{Via Antonio Baiamonti 25}, \city{Rome}, \postcode{00195}, \country{Italy}}}

\affil[2]{\orgdiv{Istituto di Analisi dei Sistemi ed Informatica ``A. Ruberti''}, \orgname{Consiglio Nazionale delle Ricerche}, \orgaddress{\street{Via dei Taurini 19}, \city{Rome}, \postcode{00185}, \country{Italy}}}

\affil[3]{\orgdiv{Dipartimento di Ingegneria Informatica, Automatica e Gestionale ``A. Ruberti''}, \orgname{Sapienza Università di Roma}, \orgaddress{\street{Via Ariosto 25}, \city{Rome}, \postcode{00185}, \country{Italy}}}


\abstract{Financial distress of municipalities, although comparable to bankruptcy of private companies, has a far more serious impact on the well-being of communities. For this reason, it is essential to detect deficits as soon as possible. Predicting financial distress in municipalities can be a complex task, as it involves understanding a wide range of factors that can affect a municipality's financial health. In this paper, we evaluate machine learning models to predict financial distress in Italian municipalities. Accounting judiciary experts have specialized knowledge and experience in evaluating the financial performance, and they use a range of indicators to make their assessments. By incorporating these indicators in the feature extraction process, we can ensure that the model is taking into account a wide range of information that is relevant to the financial health of municipalities. The results of this study indicate that using machine learning models in combination with the knowledge of accounting judiciary experts can aid in the early detection of financial distress, leading to better outcomes for the communities.}




\maketitle

\section{Introduction}
\label{introduction}
Financial distress of local governments and municipalities can have a significant impact on the delivery of essential services and the well-being of communities. The inability of local governments to meet their financial obligations can result in reduced services, higher taxes, and even bankruptcy. The bankruptcy of Detroit in 2013, being the largest US city to file, is the most well-known case. However, similar defaults by local authorities have also been observed in Europe, such as in Italy \citep{gregori2019determinants}. The financial health of local governments in Italy is of particular concern due to the ongoing fiscal challenges faced by the country. The 7904 municipalities are an important part of the Italian country, not only due of the millennia-old origins of some of them, but also for the administrative function they cover in local government areas. The Constitution permits regulatory, organizational, administrative, and financial autonomy, which, however, must be balanced by a rigorous control activity that is carried out by \textit{Corte dei conti}, the Italian supreme audit institution. Audits push municipalities toward proper management of public resources but, at the same time, must bring out financial crises as soon as possible. A state of financial distress exists if the local authority cannot guarantee indispensable functions and services, or if there are claims against the local authority from third parties that cannot be validly met. While in the private sector insolvency leads to bankruptcy and cessation of activities, in Italy, as in most developed countries, this cannot happen for local governments, which do not have to stop their activities. 
To respond to the growing financial problems of local administrations, Law n. 66 of 1989 introduced a special procedure of insolvency for municipalities. Since its introduction, the financial distress procedure for Italian local governments has undergone numerous regulatory changes, from the introduction of preventive statuses such as pre-financial distress to the provision of penalties for guilty administrators. Since the 1990s, the number of municipalities in financial distress gradually decreased, but in the last decade, it increased again. In 2021, many local authorities experienced more than one recourse to insolvency procedures: in total, nearly 5\% of Italian local authorities are in financial distress or pre-distress, with a clear predominance of municipalities in the south of Italy \citep{antulov2021predicting}.

Under the current legislation, the identification of municipalities in structural deficit is done by analyzing a fixed number of financial and economic indices. The main limitation of this approach is that it can only detect financial distress retrospectively and does not provide early warning of structural deficits \citep{markose2021early}. 
Artificial intelligence (AI), instead, can play a valuable role in predicting financial distress by analyzing various data sources and identifying early warning signs. 
This type of decision support system could help the activity of the accounting judiciary in several ways. 
For example, the predictive model could be used to carry out an initial classification of financial data from municipalities, distinguishing each year between municipalities that do not present critical issues and those that warrant further investigation. 
It is important to note that while AI can be a valuable tool, it should not replace the work of financial oversight institutions. Municipalities should utilize AI as an aid to decision-making, combining it with the knowledge and experience of financial experts to make well-informed decisions and policies.

\section{Related work}
Predicting the financial distress of public administrations is closely related to the task of bankruptcy prediction, which has been widely studied in the literature \citep{lin2011machine, sun2014predicting, son2019data, huang2022improving}. A variety of machine learning algorithms and techniques can be used for this task, such as ensemble learning \citep{cho2010hybrid, chen2020ensemble}, logistic regression \citep{hauser2011predicting}, support vector machines \citep{min2005bankruptcy, yang2011using}, genetic algorithms \citep{gordini2014genetic}, and artificial neural networks \citep{charalambous2000comparative, elhoseny2022deep, abid2022new}. The best approach depends on the specific characteristics of the data and the desired level of accuracy and interoperability \citep{barboza2017machine, devi2018survey}. However, predicting financial distress in municipalities differs from predicting bankruptcy in the corporate sector in a number of ways. One key difference is the complexity of local government operations and the potential political factors that may influence financial decisions. Municipalities have different financial reporting requirements and have access to different types of funding, which can make it more difficult to identify potential financial distress. Another difference is that municipalities have a different set of stakeholders, such as citizens, local businesses and other municipalities, which may have different priorities and concerns than the shareholders of a corporation. This can make it more challenging to predict financial distress in municipalities, as the financial well-being of a municipality may not always be reflected in its financial statements \citep{cohen2012assessing}. Additionally, municipalities have different types of financial obligations, such as long-term debt, pensions, and other post-employment benefits. These financial obligations may not be present in the corporate sector and may require specialized knowledge for accurate prediction.
Recently, there has been a growing body of literature focusing on the financial distress of local municipalities \citep{galariotis2016novel, gregori2019determinants}. These studies aim to determine the factors that influence the financial performance of local governments, such as the relationship between different levels of administration, the availability of organizational resources, the age and experience of politicians, the wealth of the local citizens, and changes in the financial environment. The literature on predicting financial distress has been primarily focused on developing statistical and machine learning models using financial and demographic data \citep{alaminos2018data, antulov2021predicting}. However, a major limitation of these studies is that they do not extract specialized features from financial and econometric indicators for training predictive models. As a result, these models may not be sufficiently reliable to support the decision-making process.

It is widely known that involving domain experts in the process of building a predictive model can help ensure that the model is accurate and relevant to the domain, and can provide valuable insights and guidance throughout the process \citep{mcgovern1989incorporating, webb1996integrating}.
In this paper, thanks to the information provided by \textit{Corte dei conti}, we build a classification model for early detection of financial crisis of local governments in Italy. The classification task is challenging due to the small number of bankruptcy cases on which learning is possible. In such cases, a machine learning model might be biased towards the majority class and not perform well on the minority class. Overall, the identification of a wide range of factors that affect a municipality’s financial health is fundamental for building a reliable prediction model.
The main contributions of this paper are:
\begin{enumerate}
    \item By focusing on predicting financial distress in municipalities we address a novel and under-explored area of financial prediction, as opposed to the more common task of bankruptcy prediction in private or corporate contexts. 
    \item Our methodology includes the use of domain-specific financial indicators and incorporates expert knowledge to enhance the predictions. Specifically, feature extraction is assisted by experts that are responsible for the financial control and auditing of public administrations. 
    \item We provide insights on the contribution of each feature used in the model to predict the outcome of financial distress, as well as its explainability.
\end{enumerate}
Results show that our model minimizes the number of false positives and maximizes predictions with high true positive rate (i.e., municipalities in crisis), thus it can be safely used in production to support the decision-making process. Additionally, this can be valuable for policymakers, financial authorities, and municipalities themselves as it can help them take proactive measures to prevent or mitigate financial distress.

The reminder of the paper is organized as follows. In Section \ref{features}, we describe the dataset and the adopted set of features, explaining their significance and interpretation. In Section \ref{method} we describe the methodology and the implementation details. In Section \ref{result}, we report the computational results and analyze the output of the model. Finally, Section \ref{concl} concludes the paper.

\section{Data description and analysis}
\label{features}
The data analyzed in this paper are collected from two distinct sources. We first analyze the historical archive provided by the University Ca' Foscari of Venice, which includes information about municipalities that have experienced financial crisis between the years 1989 and 2020. Then, we merge this dataset with data from \textit{Corte dei conti} containing all information, documentation, and financial data related to local governments over the period 2016-2020. As a result, our analysis covers a five-year period from 2016 to 2020 and includes information of 7904 Italian municipalities. The final dataset includes a total of 39520 instances, with only 416 of them being municipalities in financial distress in a given year. Note that, while the study conducted by \cite{antulov2021predicting} is based on indices provided by the Italian National Institute of Statistics and the Ministry of the Interior, the data used for the training of our model are based on those provided by legislation and used by the accounting judiciary to determine whether a municipality is in financial crisis or not. 
Accounting judiciary experts have specialized knowledge and experience in evaluating the financial performance of municipalities, and they use a range of financial and general indicators to make their assessments. By incorporating these indicators into the feature extraction process, we can ensure that our model is taking into account a wide range of information that is relevant to the financial health of municipalities \citep{lin2019feature, abid2022new}. In the following, we provide a comprehensive overview of the selected features, explaining their significance and interpretation based on the experts' domain knowledge. 



\begin{description}
    \item[Demographic category]
Classification of the municipality into 12 demographic categories according to Italian law. Demographic category is encoded as a categorical variable and is used to classify the municipality on the basis of its number of residents, see Table \ref{table:demo}.
\begin{table}[ht]
\small
\centering
    \begin{tabular}{|c | c|} 
    \hline
    Demographic category & Number of residents \\
    \hline\hline
     I & \(p < 500\) \\ 
    \hline
    II &  \(500 \leq p < 1000\) \\
    \hline
    III &  \(1000 \leq p < 2000\) \\
    \hline
    IV &  \(2000 \leq p < 3000\) \\
    \hline
    V &  \(3000 \leq p < 5000\) \\
    \hline
    VI &  \(5000 \leq p < 10000\) \\
    \hline
    VII &  \(10000 \leq p < 20000\) \\
    \hline
    VIII &  \(20000 \leq p < 60000\) \\
    \hline
    IX &  \(60000 \leq p < 100000\) \\
    \hline
    X &  \(100000 \leq p < 250000\) \\
    \hline
    XI &  \(250000 \leq p < 500000\) \\
    \hline
    XII & \(p \geq 500000\) \\ 
    \hline
    \end{tabular}
    \caption{Demographic categories and number of residents $p$ for each category.}
    \label{table:demo}
\end{table}

    \item[Geographical area] Italy is divided into 5 geographical macro-areas: north-west, north-east, center, south and islands. In Figure \ref{fig:geoarea} we report the number of municipalities in financial distress for each geographical area. This feature is encoded as a categorical variable. In our dataset, there is a clear predominance of municipalities in regions of the south of Italy. This evidence highlights the gap between Northern and Southern Italy in terms of financial distresses.
    \begin{figure}[!ht]
    \centering
        \includegraphics[scale=0.47]{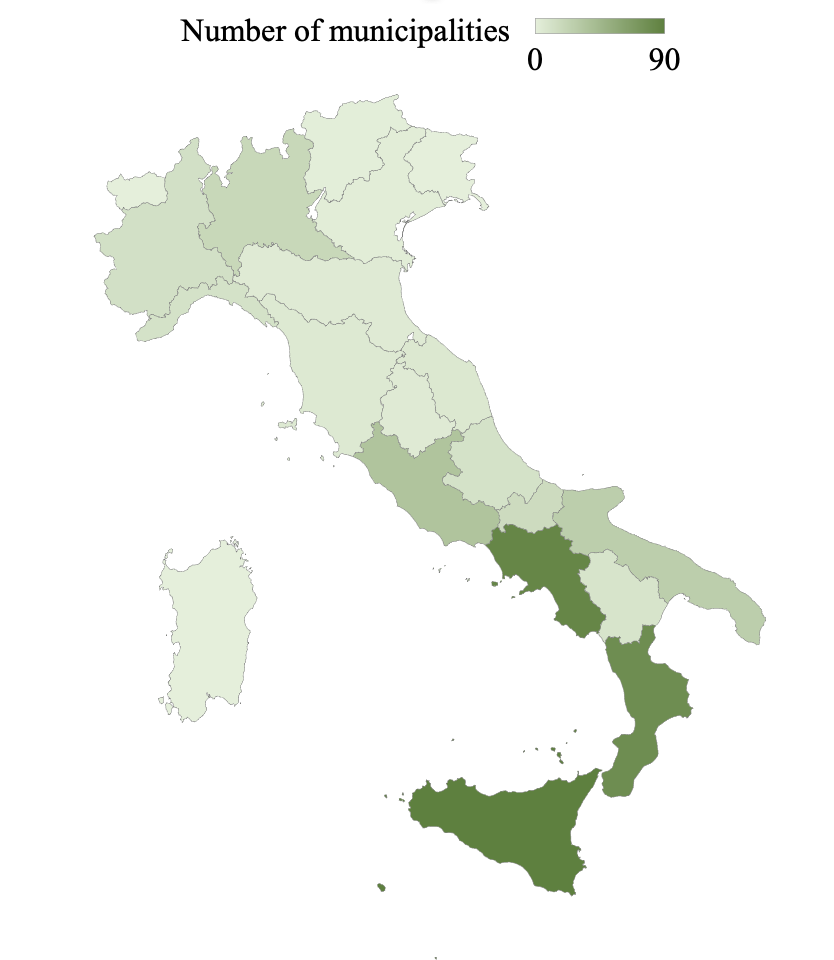}
        \caption{Municipalities in financial crisis for each geographical area.}
        \label{fig:geoarea}
    \end{figure}
    \newline
    \item[Incidence of investment and capital expenditure] It measures the portion of expenditure that the municipality decides to invest in long-term projects for its development. It is calculated as the average percentage over the last three years: the higher the value, the higher the municipality's propensity to invest.
    \newline
    \item[Financial autonomy degree] It is a measure of a municipality's ability to fulfill its financial requirements independently. It is calculated as a percentage, representing the proportion of the municipality's financial requirements that are met through its own resources. A high value indicates that the municipality is able to meet a large proportion of its financial requirements through its own resources, which indicates a high degree of financial autonomy. This could be a sign of a strong financial position and good governance. On the other hand, a low value indicates that the municipality is not able to meet a large proportion of its financial requirements through its own resources, which indicates a low degree of financial autonomy. This could be an indication of a weak financial position and poor governance.
    \newline
    \item[Indebtedness per capita] It is a measure of a municipality's financial health and solvency. It is calculated by dividing the municipality's total debt by its resident population. A high value of indicates that the municipality has a large amount of debt relative to its population, which could be a sign of financial strain. On the other hand, a low value indicates that the municipality has a relatively low amount of debt and may be in a stronger financial position.
    \newline
    \item[Total investment financed by debt] It is a measure of a municipality's use of debt to finance its investments. It is calculated as the percentage of the municipality's total investments that are financed through debt. A high value indicates that the municipality is heavily relying on debt to finance its investments, which could be a sign of financial strain. On the other hand, a low value indicates that the municipality is using more equity or other forms of financing to fund its investments, which may be a sign of a stronger financial position.
    \newline
    \item[Rigid expenditure] It a measure of a municipality's flexibility in managing its operating expenses. It is calculated as a percentage and represents the proportion of the municipality's operating expenses that are fixed and cannot be easily reduced in the short term. A high value indicates that the municipality has a high proportion of fixed expenses, which reduces its ability to take action to decrease operating expenses in the short term. This could be a sign of financial strain, as the municipality may have less ability to respond to unexpected changes in revenue or other financial pressures. On the other hand, a low value indicates that the municipality has a low proportion of fixed expenses, and greater flexibility to reduce operating expenses if needed, which is a sign of greater financial stability.
    \newline
    \item[Expense management speed] It measures the municipality's efficiency in managing its expenses. It is calculated as the percentage of the municipality's commitments that are paid within the current year. A high value indicates that the municipality is paying its commitments in a timely manner and has an efficient administration. This could be a sign of good financial management and good governance. On the other hand, a low value of expense management speed indicates that the municipality is not paying its commitments in a timely manner and the administration may not be as efficient. This could be an indication of poor financial management and governance.
    \newline
    \item[Collecting capacity] It is a measure of a municipality's ability to collect taxes as planned from the beginning of the year. It is calculated as a percentage, representing the proportion of the municipality's expected tax collection that is actually collected. A high value indicates that the municipality is able to collect a high portion of the taxes that it planned to collect. This could be a sign of good tax administration and compliance by taxpayers. On the other hand, a low value indicates that the municipality is not able to collect a high portion of the taxes that it planned to collect. This could be an indication of poor tax administration and non-compliance by taxpayers.
    \newline
    \item[Extra-budgetary debts] It represents payment obligations that a municipality has contracted with third parties without specific programmed financial coverage. These debts are not included in the municipality's budget, and they can be a sign of financial strain. A high value indicates that the municipality has a large amount of payment obligations to third parties that are not included in the budget, which can be a sign of poor financial management and governance. On the other hand, a low value of indicates that the municipality has a relatively low amount of payment obligations to third parties that are not included in the budget, which may be a sign of better financial management and governance.
    \newline
    \item[Bankruptcy risk] It is a measure of a municipality's likelihood of facing financial distress or being unable to meet its financial obligations. We classify it into five levels of severity, ranging from ``low risk'' (level 1) to ``high risk'' (level 5), based on the municipality's past financial performance. A high risk level indicates that the municipality has a high likelihood of facing financial distress or being unable to meet its financial obligations. This could be a sign of poor financial management, high levels of debt, or a weak economy. On the other hand, a low risk level indicates that the municipality has a low likelihood of facing financial distress or being unable to meet its financial obligations. This could be a sign of strong financial management, low levels of debt, or a strong economy.
    \newline\newline
    In Table \ref{tab:risk}, we report the adopted severity risk scale. This feature enables us to incorporate information on the duration of financial distress. For instance, if a municipality has a history of multiple bankruptcies (high risk, according to our classification), it is more likely to experience financial distress again than a municipality that has never gone bankrupt. Some local governments, due to their geographical and social context, may be more susceptible to structural deficits and, as a result, more prone to financial distress situations. In Figure \ref{fig:risk&dist}, we show the bankruptcy risk for the 416 municipalities in financial crisis.

    \begin{table}[!ht]
    \small
    \centering
    \begin{tabular}{|l|l|}
        \hline
         Bankruptcy risk value & Motivation  \\
        \hline\hline
        \begin{tabular}{l}
            1 (low)
        \end{tabular} &
        \begin{tabular}{l}
            The municipality has experienced \\ pre-distress at most once.
        \end{tabular} \\
        \hline
        \begin{tabular}{l}
            2 (medium-low)
        \end{tabular} &
        \begin{tabular}{l}
            The municipality has experienced \\ pre-distress multiple times.
        \end{tabular} \\
        \hline
        \begin{tabular}{l}
            3 (medium)
        \end{tabular} &
        \begin{tabular}{l}
            The municipality has only experienced \\ bankruptcy once.
        \end{tabular} \\
        \hline
        \begin{tabular}{l}
            4 (medium-high)
        \end{tabular} &
        \begin{tabular}{l}
            The municipality has first experienced \\
            pre-distress and then bankruptcy.
        \end{tabular} \\
        \hline
        \begin{tabular}{l}
            5 (high)
        \end{tabular} &
        \begin{tabular}{l}
            The municipality has experienced \\
            bankruptcy multiple times.
        \end{tabular} \\
        \hline
\end{tabular}
\caption{In this table we describe how the risk level is assigned to each municipality.}
    \label{tab:risk}
\end{table}
    
    \begin{figure}[!ht]
    \centering
    \includegraphics[scale=0.55]{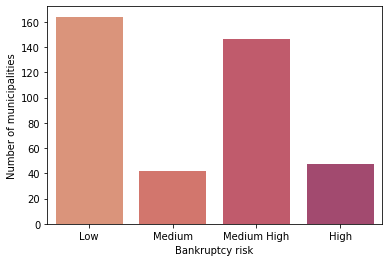}
    \caption{Bankruptcy risk value for municipalities in financial distress.}
    \label{fig:risk&dist}
    \end{figure}
    \item[Off-balance sheet debts] They represent an obligation towards third parties for the payment of a specific amount of money, assumed in violation of the justifiable rules that govern the expenditure procedures of local authorities. The issue of off-balance sheet debts is one of the most widespread pathologies in municipalities that can be configured as the main cause of heavy financial imbalances to be restored. We consider a binary variable indicating whether the municipality has off-balance sheet debts or not.  
\end{description}

To identify patterns or conditions that are characteristic of financial distress, we also incorporate lagged features in the prediction model. The concept behind this is that by observing how a particular feature has changed from one time to the next, the model can better capture the underlying trends in the data. Specifically, we compute the difference between a feature at year $t$ and at year $t-1$, and then introduce a new variable to represent this change. 
Lagged predictors are computed for ``expense management speed'', ``rigid expenditure'', ``total investment financed by debt'', ``financial autonomy degree'', ``collecting capacity, and ``indebtedness per capita''.

\begin{figure}[!ht]
    \centering
    \includegraphics[scale=0.6]{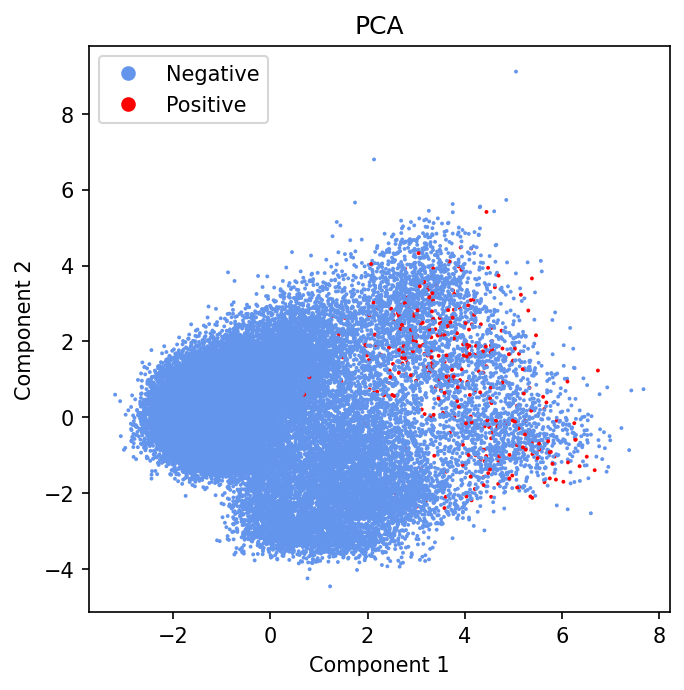}
    \caption{Distribution of the municipalities in the PCA space.}
    \label{fig:pca_plot}
\end{figure}

We use the Principal Component Analysis (PCA) to visualize the feature-spaces of municipalities. The results are shown in Figure \ref{fig:pca_plot}, where we plot the first two principal components against each other. In this plot, the first and the second principal component explain 62.5\% of the total variation of the data. The points are individual municipalities, and their color indicates their binary label (blue for the negative label and red for the positive label). The municipalities that are not in financial distress (blue) are mostly concentrated in the bottom left corner, while the municipalities in financial distress (red) are mostly concentrated in the top right. Furthermore, the PCA plot presents a clear visualization of several clusters in the data, which suggests that there may be distinct groups of municipalities with similar characteristics. However, in this paper, we have chosen to focus on a supervised classification approach using the available labels rather than performing cluster analysis \citep{du2016two, piccialli2022sos}. This decision was made because the labels provide a direct way to evaluate the performance of the model and make predictions. 

\section{Methodology}
\label{method}
In this section, we illustrate the proposed methodology for predicting financial distress of municipalities. We first formalize the prediction task and then we describe the experimental setup used to train the machine learning models.

\subsection{Modeling}
We are given a set of 7094 municipalities and the corresponding set $T = \{2016, \dots, 2020\}$ of 5 years. Each municipality is characterized a the set of features $x_t$ describing its financial, institutional and economic context at time $t \in T$, as reported in Section \ref{features}. Therefore, an object of our dataset $M$ is a municipality at time $t$ and we want to classify each of them into one of two classes, which we will denote as ``positive'' and ``negative''. More precisely, every municipality $x_t$ has associated the target binary variable $y_t$, that takes the value 1 (positive sample) if bankruptcy has occurred at time $t$ and value 0 (negative sample) otherwise. Thus, our goal is to learn a classification function $f : M \rightarrow [0, 1]$ that takes as input a municipality $x_t \in M$ and outputs the class probability $\hat{y}_t = f(x_t)$ for it. Once the classifier has been trained, we can use the function $f$ to make predictions on new municipalities by computing the associated probability and classifying them by choosing a discrimination threshold. We remark that our dataset is highly unbalanced since the number of municipalities in financial distress is much smaller than the number of healthy ones. This poses a challenge for training a binary classification model, as the model may tend to predict the majority class for every municipality, leading to poor performance on the minority class. To address this issue, we incorporate class weights in binary classification models. The idea behind using class weights is to assign higher importance to objects from the minority class (municipalities in financial distress), which will lead the model to pay more attention to them during training. 
We divide the data in training and testing sets, with 80\% of the data used for training and the remaining 20\% used for testing. We calculate class weights as the inverse of the class frequencies in the training data. 

\subsection{Experimental setup}
Before training machine learning models, we follow standard pre-processing steps. First, we perform data cleaning by identifying and correcting errors in the data. Next, we transform the data by standardizing numeric variables and converting categorical variables into a numerical representation using one-hot encoding. We consider four different models for classification problems: Logistic regression, Support Vector Machine (SVM), Random forest, and Extreme Gradient Boosting (XGBoost). Logistic regression is a widely known statistical method for binary classification. SVM is a powerful algorithm for classification and regression, particularly in high-dimensional spaces \citep{cortes1995support}. Random Forest, an ensemble method, combines multiple decision trees to improve the predictive performance of the model \citep{breiman2001random}. Lastly, XGBoost is a gradient boosting algorithm that has been shown to outperform other ensemble methods in many benchmark datasets \citep{chen2016xgboost}. 

To evaluate the performance of the models we use the F1 score, i.e., the harmonic mean of precision and recall, where precision is the number of true positive predictions divided by the sum of true positive and false positive predictions, and recall is the number of true positive predictions divided by the sum of true positive and false negative predictions. To obtain a reliable estimate of the performance we employ a 5-fold cross-validation strategy. The training set is randomly split into 5 equal folds, with each fold representing 20\% of the total. We train the model on 4 folds and test it on the remaining fold. This process is repeated for each of the 5 folds, such that each fold served as the test set once. The results are then averaged to obtain a final estimate of the model's performance. To ensure a fair comparison, we stratified the folds so that the distribution of classes within each fold is the same as the overall distribution in the dataset.
We employ a grid search method to tune the hyperparameters of the classifiers. For the logistic regression, the hyperparameters are the regularization type \{``L1'', ``L2''\} and the penalty parameter \{0.1, 0.5, 1, 5, 10\}. For SVM, the following hyperparameters are tuned: kernel type \{``linear'', ``radial basis function''\}, regularization parameter C \{0.1, 0.5, 1, 5, 10\}, and gamma parameter \{0.001, 0.01, 0.1\}. The hyperparameters for Random forest are the number of trees in the forest \{100, 200, 300, 400\}, the maximum depth of the trees \{3, 5, 7\}, and the minimum number of samples required to split an internal node \{2, 5, 10\}. Finally, the hyperparameters for XGBoost are the maximum depth of the trees \{3, 5, 7\}, the learning rate \{0.1, 0.01, 0.001\}, and the number of estimators \{100, 200, 300, 400\}. The grid search evaluates all possible combinations of hyperparameters for each model and returns the model that results in the highest average  ``macro'' F1 score, i.e., the average F1 score across all classes, treating each class as equally important. Finally, the model with the highest F1 score is retrained on the whole training set and tested on the hold-out test set.

\section{Results and discussion}
\label{result}
\begin{figure}[!ht]
    \centering
    \includegraphics[scale=0.65]{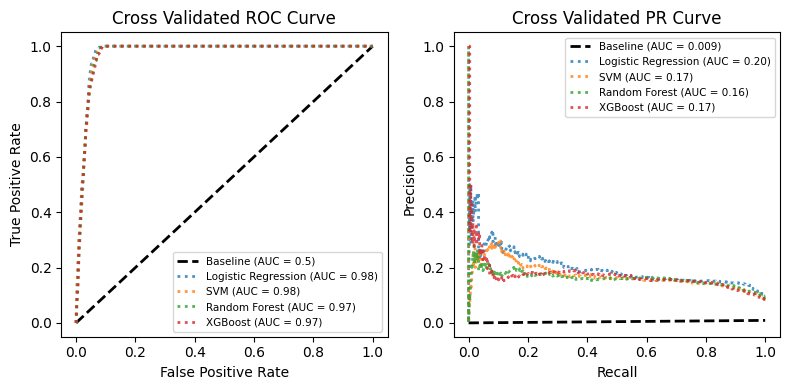}
    \caption{Cross validated ROC and PR curves for the considered models.}
    \label{fig:roc_pr}
\end{figure}

For logistic regression, the optimal parameters are an L2 regularization with a penalty of 5. For random forest, the optimal parameters are 100 estimators, a maximum depth of 7 and the minimum number of samples required to split an internal node equal to 5. For SVM, the optimal parameters are a linear kernel and a penalty parameter of 1. Finally, for XGBoost, the optimal parameters are 100 estimators, a maximum depth of 5 and learning rate of 0.1.
In Figure \ref{fig:roc_pr}, we show the average Receiver Operating Characteristic (ROC) curve and Precision-Recall (PR) curve within the cross-validation procedure. Both ROC and PR curves illustrate the diagnostic ability of a binary classifier as its discrimination threshold is varied. The ROC curve shows the true positive rate against the false positive rate at different threshold settings \citep{bradley1997use}. The area under the ROC curve (AUC-ROC) summarizes the performance of the classifier. On the other hand, the PR curve displays the precision and recall scores of a classifier at different threshold settings. Similarly to the AUC-ROC, the area under the PR curve (AUC-PR) provides a single scalar value that summarizes the performance of the model. Both AUC-ROC and AUC-PR can range from 0 to 1, with a score of 1 indicating a perfect classifier \citep{davis2006relationship}. Note that, the ``baseline curve'' in the PR curve plot is a horizontal line with height equal to the number of positive examples over the total number of training data, i.e. the proportion of positive examples in our data.

The ROC curve shows that all four methods perform similarly. However, a closer examination of the PR curve reveals that there are differences between the methods in terms of their ability to correctly classify positive instances. Given the highly unbalanced nature of the dataset, we can conclude that the PR curve provides a more nuanced and accurate assessment of the performance of each method, compared to the ROC curve. 
Indeed, we chose the logistic regression as our final model due to its simplicity, interpretability, and computational efficiency. This decision is based on the principle of parsimony, which states that the simplest explanation that fits the data is preferred \citep{guyon2010model}. Additionally, logistic regression is a widely used and well-established method in the related literature and has been shown to perform well on a variety of of bankruptcy prediction tasks, see, e.g., \cite{barboza2017machine} and \cite{son2019data}.


The confusion matrix of the logistic regression computed on the test set is shown in Figure \ref{fig:conf_matrix}. The entries in the matrix correspond to the number of true positives (TP), false negatives (FN), false positives (FP), and true negatives (TN). Notably, the classifier correctly labels all municipalities in financial distress, and this is shown in the confusion matrix by the absence of false negatives. Conversely, the confusion matrix displays 420 false positives. However, we remark that, when predicting financial distress in municipalities, having a small number of false negatives is more important than having a small number of false positives because a false negative means that the classifier is not detecting a municipality that is actually in financial distress. This could lead to the municipality not receiving the necessary support or intervention, which could have serious consequences. On the other hand, a false positive means that the classifier is incorrectly identifying a municipality as being in financial distress when it is not. While this could lead to unnecessary costs or resources being allocated, it is less severe than not detecting a municipality that is actually in financial distress. We now examine false positives in more detail.

\begin{figure}[!ht]
\label{fig:conf_matrix}
    \begin{center}
    {

        \offinterlineskip

        \raisebox{-5.5cm}[0pt][0pt]{
            \parbox[c][5pt][c]{1cm}{\hspace{-3.0cm}\rot{\textbf{True Label}}\\[85pt]}}\par

        \hspace*{1cm}\MyHBox[\dimexpr2.4cm+6\fboxsep\relax]{Predicted Label}\par

        \hspace*{1cm}\MyHBox{1}\MyHBox{0}\par

        \MyTBox{{1}}{{TP \\ \textbf{67}}}{{FN \\ \textbf{0}}}

        \MyTBox{{0}}{{FP \\ \textbf{420}}}{{TN} \\ \textbf{7215}}

    }
\end{center}
\caption{Confusion matrix of the logistic regression on the test set.}
\end{figure}

False positives can also include municipalities that are showing signs of financial distress but are not yet classified as such. Such municipalities may belong to the so-called ``grey area'', a term used in the literature to refer to situations or entities that are uncertain or difficult to classify \citep{cortes2016learning}. 
Specifically, these municipalities might have some financial indicators that are concerning, such as high debt levels or budget deficits, but they have not yet reached the point of being officially classified as being in financial distress. These ``grey areas'' can be considered as a ``warning sign'' or ``early warning'' of potential financial distress, and it is important to identify them early in order to take appropriate actions and prevent a full-blown financial crisis.

To carry out the false positives analysis, we make predictions on 2016 data of the test set where there are 1449 municipalities with 13 being in financial crisis. The model produces 13 true positives and 117 false positives. Of these 117 municipalities, we evaluate how many of them would experience financial crisis in the years following 2016 (2017, 2018, 2019 and 2020). It turns out that a significant number of these false positives, specifically 72 municipalities or 61.5\% of the total, would actually go into financial distress in the following four years. This highlights the importance of analyzing false positives in predictive models for financial distress and lays the foundation for a model that can provide medium, short, or long-range predictions. Furthermore, this early identification can assist officials to take appropriate actions, such as implementing financial management measures, and also help in monitoring the situation more closely, to mitigate the impact of financial distress.

\begin{figure}[!ht]
    \centering
    \includegraphics[scale=0.6]{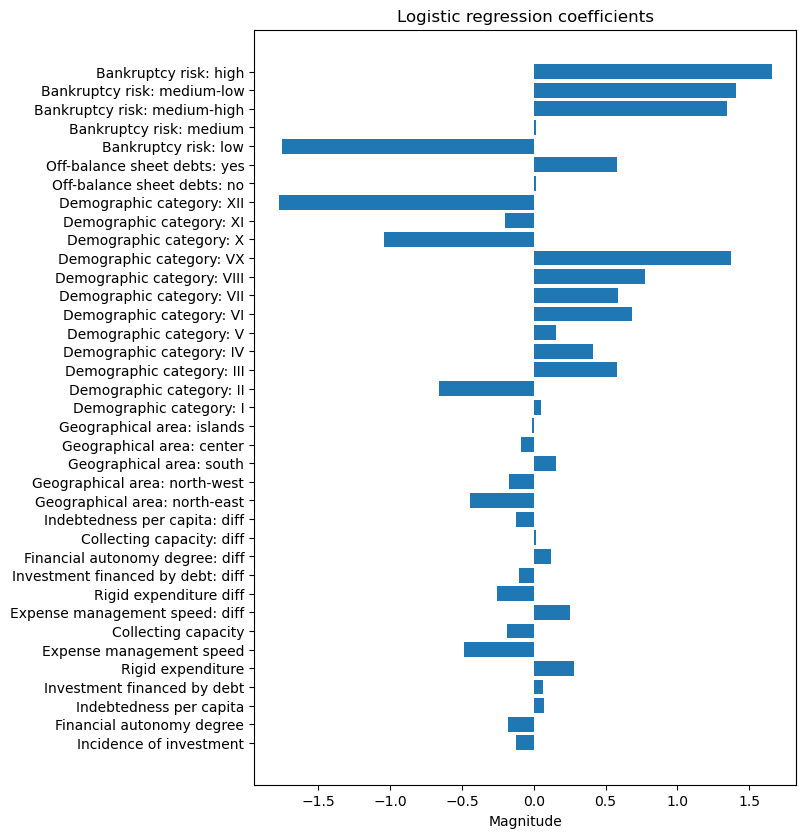}
    \caption{Coefficients of the logistic regression for each feature.}
    \label{fig:coeff}
\end{figure}

Finally, we conclude our analysis by showing the coefficients of the logistic regression model. In addition to out-of-sample predictions, the coefficients of logistic regression also provide information on the impact of each feature in explaining the binary outcome of financial distress. A positive coefficient means that as the predictor value increases, the probability of financial distress also increases, while a negative coefficient means that as the predictor value increases, the probability of financial distress decreases. The magnitude of the coefficient represents the strength of the relationship between the feature and the financial distress. We can easily interpret the coefficients in Figure \ref{fig:coeff} on the basis of the knowledge and understanding of the hand-crafted features described in Section \ref{features}. For example, the one-hot encoded categorical feature ``bankruptcy risk'' has large coefficients, which means that as the bankruptcy risk reaches the highest severity level, the probability of financial distress also increases. Feature ``off-balance sheet debts'' indicates whether the municipality has off-balance sheet debts or not. The positive coefficient in the model indicates that as the off-balance sheet debts increase, the probability of financial distress increases. Notably, the coefficients associated to ``demographic category'' show that municipalities with the largest number of residents are more likely to experience financial distress. One possible explanation of this phenomenon is that larger municipalities may have more complex financial systems and higher levels of debt, making it harder to manage finances effectively. These can include rising demands for public services and infrastructure, the costs of which can outpace the growth of tax revenues. Speaking of taxes, the coefficient of ``collecting capacity'' has a negative value indicating that a high value of the corresponding feature pushes the classification more towards the negative class. The same observation also holds for the coefficient of ``expense management speed'' where a high value of the feature indicates that the municipality is paying its commitments in a timely manner and has an efficient administration. Finally, the coefficients associated to the dummy variables for the feature ``geographical area'' highlight that municipalities in Northern Italy are less likely to experience financial distress. 


\section{Conclusions and future works}
\label{concl}
In this paper, we evaluated machine learning models for predicting financial distress of Italian municipalities. A preliminary stage of feature engineering was necessary to determine domain-specific features for the prediction model. This stage was enhanced by specialized knowledge provided by experts of \textit{Corte dei conti}, the Italian supreme audit institution. By incorporating information used by the accounting judiciary in the feature extraction process, we provided an accurate assessment of a municipality's financial health. By comparing standard classification algorithms, the logistic regression achieved maximum sensitivity, that is, the ability to correctly classify positive cases. We have also illustrated the results of the analysis conducted on false positives, showing that more than half of the mislabeled municipalities would actually experience financial crisis in the next 4 years. Finally, thanks to the coefficients of the logistic regression we evaluated the impact of the features used in the model in explaining the outcome of financial distress.
As a future research direction, it may be beneficial to explore clustering techniques to identify similarities between municipalities in financial distress.
Specifically, a semi-supervised approach, where background knowledge provided by domain experts is incorporated into the clustering process, may be considered \citep{karlos2016effectiveness, piccialli2022exact}. 
Furthermore, exploring hierarchical models as described by \cite{mancuso2021machine} may also be beneficial. Given that a country's administrative-territorial structure is often divided into hierarchically subordinate parts, predicting financial distress in a municipality should take into account not only internal factors, but also the regulatory impact of regional authorities. Using hierarchical models allows for the examination of the influence of different levels of public administration hierarchy and neighboring territories on the indicators of socio-economic condition of municipal entities.

\section*{Acknowledgements}
This work was supported by \textit{Corte dei conti}, the Italian supreme audit institution. In this regard, heartfelt thanks go to the \textit{Sezione delle autonomie} and its presidents Francesco Petronio and Fabio Viola for granting access to the datamart on local government finance. We would like to thank the whole ICT Department of \textit{Corte dei conti}, with a particular mention for the magistrate in charge of ICT, Carlo Mancinelli, CIO Luca Attias and IT managers Francesca Tondi and Leandro Gelasi. We thank magistrates Marcello Degni and Giampiero Maria Gallo for their support for feature engineering. We also thank Ca' Foscari University of Venice which made available the data used to build the target feature. Many useful tips were provided by magistrate Gerardo De Marco.

\section*{Statements and Declarations}
\subsection*{Funding}
The authors declare that no funds, grants, or other support were received during the preparation of this manuscript.
\subsection*{Competing Interests}
The authors have no financial or non-financial interests to disclose.
\subsection*{Data Availability}
The data that support the findings of this study are available from the corresponding author upon request.

\bibliography{sn-bibliography}

\begin{thebibliography}{36}
\expandafter\ifx\csname natexlab\endcsname\relax\def\natexlab#1{#1}\fi
\expandafter\ifx\csname url\endcsname\relax
  \def\url#1{{\tt #1}}\fi
\expandafter\ifx\csname urlprefix\endcsname\relax\def\urlprefix{URL }\fi

\bibitem[{Abid et~al.(2022)Abid, Ayadi, Guesmi, \& Mkaouar}]{abid2022new}
Abid, I., Ayadi, R., Guesmi, K., \& Mkaouar, F. (2022).
\newblock A new approach to deal with variable selection in neural networks: an
  application to bankruptcy prediction.
\newblock {\em Annals of Operations Research\/}, pp. 1--19.

\bibitem[{Alaminos et~al.(2018)Alaminos, Fern{\'a}ndez, Garc{\'\i}a, \&
  Fern{\'a}ndez}]{alaminos2018data}
Alaminos, D., Fern{\'a}ndez, S.~M., Garc{\'\i}a, F., \& Fern{\'a}ndez, M.~A.
  (2018).
\newblock Data mining for municipal financial distress prediction.
\newblock In {\em Industrial Conference on Data Mining\/}, pp. 296--308.
  Springer.

\bibitem[{Antulov-Fantulin et~al.(2021)Antulov-Fantulin, Lagravinese, \&
  Resce}]{antulov2021predicting}
Antulov-Fantulin, N., Lagravinese, R., \& Resce, G. (2021).
\newblock Predicting bankruptcy of local government: A machine learning
  approach.
\newblock {\em Journal of Economic Behavior \& Organization\/}, {\em 183\/},
  681--699.

\bibitem[{Barboza et~al.(2017)Barboza, Kimura, \& Altman}]{barboza2017machine}
Barboza, F., Kimura, H., \& Altman, E. (2017).
\newblock Machine learning models and bankruptcy prediction.
\newblock {\em Expert Systems with Applications\/}, {\em 83\/}, 405--417.

\bibitem[{Bradley(1997)}]{bradley1997use}
Bradley, A.~P. (1997).
\newblock The use of the area under the roc curve in the evaluation of machine
  learning algorithms.
\newblock {\em Pattern Recognition\/}, {\em 30\/}(7), 1145--1159.

\bibitem[{Breiman(2001)}]{breiman2001random}
Breiman, L. (2001).
\newblock Random forests.
\newblock {\em Machine Learning\/}, {\em 45\/}(1), 5--32.

\bibitem[{Charalambous et~al.(2000)Charalambous, Charitou, \&
  Kaourou}]{charalambous2000comparative}
Charalambous, C., Charitou, A., \& Kaourou, F. (2000).
\newblock Comparative analysis of artificial neural network models: Application
  in bankruptcy prediction.
\newblock {\em Annals of Operations Research\/}, {\em 99\/}(1-4), 403--425.

\bibitem[{Chen \& Guestrin(2016)}]{chen2016xgboost}
Chen, T., \& Guestrin, C. (2016).
\newblock Xgboost: A scalable tree boosting system.
\newblock In {\em Proceedings of the 22nd ACM SIGKDD International Conference
  on Knowledge Discovery and Data Mining\/}, KDD '16, p. 785–794. New York,
  NY, USA: Association for Computing Machinery.

\bibitem[{Chen et~al.(2020)Chen, Chen, \& Shi}]{chen2020ensemble}
Chen, Z., Chen, W., \& Shi, Y. (2020).
\newblock Ensemble learning with label proportions for bankruptcy prediction.
\newblock {\em Expert Systems with Applications\/}, {\em 146\/}, 113155.

\bibitem[{Cho et~al.(2010)Cho, Hong, \& Ha}]{cho2010hybrid}
Cho, S., Hong, H., \& Ha, B.-C. (2010).
\newblock A hybrid approach based on the combination of variable selection
  using decision trees and case-based reasoning using the mahalanobis distance:
  For bankruptcy prediction.
\newblock {\em Expert Systems with Applications\/}, {\em 37\/}(4), 3482--3488.

\bibitem[{Cohen et~al.(2012)Cohen, Doumpos, Neofytou, \&
  Zopounidis}]{cohen2012assessing}
Cohen, S., Doumpos, M., Neofytou, E., \& Zopounidis, C. (2012).
\newblock Assessing financial distress where bankruptcy is not an option: An
  alternative approach for local municipalities.
\newblock {\em European Journal of Operational Research\/}, {\em 218\/}(1),
  270--279.

\bibitem[{Cortes et~al.(2016)Cortes, DeSalvo, \& Mohri}]{cortes2016learning}
Cortes, C., DeSalvo, G., \& Mohri, M. (2016).
\newblock Learning with rejection.
\newblock In {\em International Conference on Algorithmic Learning Theory\/},
  pp. 67--82. Springer.

\bibitem[{Cortes \& Vapnik(1995)}]{cortes1995support}
Cortes, C., \& Vapnik, V. (1995).
\newblock Support-vector networks.
\newblock {\em Machine Learning\/}, {\em 20\/}(3), 273--297.

\bibitem[{Davis \& Goadrich(2006)}]{davis2006relationship}
Davis, J., \& Goadrich, M. (2006).
\newblock The relationship between precision-recall and roc curves.
\newblock In {\em Proceedings of the 23rd International Conference on Machine
  Learning\/}, pp. 233--240.

\bibitem[{Devi \& Radhika(2018)}]{devi2018survey}
Devi, S.~S., \& Radhika, Y. (2018).
\newblock A survey on machine learning and statistical techniques in bankruptcy
  prediction.
\newblock {\em International Journal of Machine Learning and Computing\/}, {\em
  8\/}(2), 133--139.

\bibitem[{du~Jardin(2016)}]{du2016two}
du~Jardin, P. (2016).
\newblock A two-stage classification technique for bankruptcy prediction.
\newblock {\em European Journal of Operational Research\/}, {\em 254\/}(1),
  236--252.

\bibitem[{Elhoseny et~al.(2022)Elhoseny, Metawa, Sztano, \&
  El-Hasnony}]{elhoseny2022deep}
Elhoseny, M., Metawa, N., Sztano, G., \& El-Hasnony, I.~M. (2022).
\newblock Deep learning-based model for financial distress prediction.
\newblock {\em Annals of Operations Research\/}, pp. 1--23.

\bibitem[{Galariotis et~al.(2016)Galariotis, Guyot, Doumpos, \&
  Zopounidis}]{galariotis2016novel}
Galariotis, E., Guyot, A., Doumpos, M., \& Zopounidis, C. (2016).
\newblock A novel multi-attribute benchmarking approach for assessing the
  financial performance of local governments: Empirical evidence from france.
\newblock {\em European Journal of Operational Research\/}, {\em 248\/}(1),
  301--317.

\bibitem[{Gordini(2014)}]{gordini2014genetic}
Gordini, N. (2014).
\newblock A genetic algorithm approach for {SME}s bankruptcy prediction:
  Empirical evidence from italy.
\newblock {\em Expert Systems with Applications\/}, {\em 41\/}(14), 6433--6445.

\bibitem[{Gregori \& Marattin(2019)}]{gregori2019determinants}
Gregori, W.~D., \& Marattin, L. (2019).
\newblock Determinants of fiscal distress in italian municipalities.
\newblock {\em Empirical Economics\/}, {\em 56\/}(4), 1269--1281.

\bibitem[{Guyon et~al.(2010)Guyon, Saffari, Dror, \& Cawley}]{guyon2010model}
Guyon, I., Saffari, A., Dror, G., \& Cawley, G. (2010).
\newblock Model selection: beyond the bayesian/frequentist divide.
\newblock {\em Journal of Machine Learning Research\/}, {\em 11\/}(1).

\bibitem[{Hauser \& Booth(2011)}]{hauser2011predicting}
Hauser, R.~P., \& Booth, D. (2011).
\newblock Predicting bankruptcy with robust logistic regression.
\newblock {\em Journal of Data Science\/}, {\em 9\/}(4), 565--584.

\bibitem[{Huang et~al.(2022)Huang, Yao, Luo, \& Li}]{huang2022improving}
Huang, B., Yao, X., Luo, Y., \& Li, J. (2022).
\newblock Improving financial distress prediction using textual sentiment of
  annual reports.
\newblock {\em Annals of Operations Research\/}, pp. 1--28.

\bibitem[{Karlos et~al.(2016)Karlos, Kotsiantis, Fazakis, \&
  Sgarbas}]{karlos2016effectiveness}
Karlos, S., Kotsiantis, S., Fazakis, N., \& Sgarbas, K. (2016).
\newblock Effectiveness of semi-supervised learning in bankruptcy prediction.
\newblock In {\em 2016 7th International Conference on Information,
  Intelligence, Systems \& Applications (IISA)\/}, pp. 1--6. IEEE.

\bibitem[{Lin et~al.(2019)Lin, Lu, \& Tsai}]{lin2019feature}
Lin, W.-C., Lu, Y.-H., \& Tsai, C.-F. (2019).
\newblock Feature selection in single and ensemble learning-based bankruptcy
  prediction models.
\newblock {\em Expert Systems\/}, {\em 36\/}(1), e12335.

\bibitem[{Lin et~al.(2011)Lin, Hu, \& Tsai}]{lin2011machine}
Lin, W.-Y., Hu, Y.-H., \& Tsai, C.-F. (2011).
\newblock Machine learning in financial crisis prediction: a survey.
\newblock {\em IEEE Transactions on Systems, Man, and Cybernetics, Part C
  (Applications and Reviews)\/}, {\em 42\/}(4), 421--436.

\bibitem[{Mancuso et~al.(2021)Mancuso, Piccialli, \&
  Sudoso}]{mancuso2021machine}
Mancuso, P., Piccialli, V., \& Sudoso, A.~M. (2021).
\newblock A machine learning approach for forecasting hierarchical time series.
\newblock {\em Expert Systems with Applications\/}, {\em 182\/}, 115102.

\bibitem[{Markose et~al.(2021)Markose, Giansante, Eterovic, \&
  Gatkowski}]{markose2021early}
Markose, S., Giansante, S., Eterovic, N.~A., \& Gatkowski, M. (2021).
\newblock Early warning of systemic risk in global banking: eigen-pair r number
  for financial contagion and market price-based methods.
\newblock {\em Annals of Operations Research\/}, pp. 1--39.

\bibitem[{McGovern \& Samson(1989)}]{mcgovern1989incorporating}
McGovern, J., \& Samson, D. (1989).
\newblock Incorporating expertise into decision analysis based {DSS}.
\newblock {\em Annals of Operations Research\/}, {\em 21\/}(1), 173--194.

\bibitem[{Min \& Lee(2005)}]{min2005bankruptcy}
Min, J.~H., \& Lee, Y.-C. (2005).
\newblock Bankruptcy prediction using support vector machine with optimal
  choice of kernel function parameters.
\newblock {\em Expert Systems with Applications\/}, {\em 28\/}(4), 603--614.

\bibitem[{Piccialli et~al.(2022{\natexlab{a}})Piccialli, {Russo Russo}, \&
  Sudoso}]{piccialli2022exact}
Piccialli, V., {Russo Russo}, A., \& Sudoso, A.~M. (2022{\natexlab{a}}).
\newblock An exact algorithm for semi-supervised minimum sum-of-squares
  clustering.
\newblock {\em Computers \& Operations Research\/}, {\em 147\/}, 105958.

\bibitem[{Piccialli et~al.(2022{\natexlab{b}})Piccialli, Sudoso, \&
  Wiegele}]{piccialli2022sos}
Piccialli, V., Sudoso, A.~M., \& Wiegele, A. (2022{\natexlab{b}}).
\newblock {SOS-SDP}: An exact solver for minimum sum-of-squares clustering.
\newblock {\em INFORMS Journal on Computing\/}, {\em 34\/}(4), 2144--2162.

\bibitem[{Son et~al.(2019)Son, Hyun, Phan, \& Hwang}]{son2019data}
Son, H., Hyun, C., Phan, D., \& Hwang, H.~J. (2019).
\newblock Data analytic approach for bankruptcy prediction.
\newblock {\em Expert Systems with Applications\/}, {\em 138\/}, 112816.

\bibitem[{Sun et~al.(2014)Sun, Li, Huang, \& He}]{sun2014predicting}
Sun, J., Li, H., Huang, Q.-H., \& He, K.-Y. (2014).
\newblock Predicting financial distress and corporate failure: A review from
  the state-of-the-art definitions, modeling, sampling, and featuring
  approaches.
\newblock {\em Knowledge-Based Systems\/}, {\em 57\/}, 41--56.

\bibitem[{Webb(1996)}]{webb1996integrating}
Webb, G.~I. (1996).
\newblock Integrating machine learning with knowledge acquisition through
  direct interaction with domain experts.
\newblock {\em Knowledge-based systems\/}, {\em 9\/}(4), 253--266.

\bibitem[{Yang et~al.(2011)Yang, You, \& Ji}]{yang2011using}
Yang, Z., You, W., \& Ji, G. (2011).
\newblock Using partial least squares and support vector machines for
  bankruptcy prediction.
\newblock {\em Expert Systems with Applications\/}, {\em 38\/}(7), 8336--8342.

\end{thebibliography}

\end{document}